\documentclass[10pt,letterpaper]{article}
\usepackage[top=0.85in,left=2.75in,footskip=0.75in,marginparwidth=2in]{geometry}
\usepackage[utf8]{inputenc}
\usepackage{cite}
\usepackage{nameref,hyperref}
\usepackage{microtype}
\DisableLigatures[f]{encoding = *, family = * }
\raggedright
\usepackage{changepage}
\usepackage[aboveskip=1pt,labelfont=bf,labelsep=period,singlelinecheck=off]{caption}
\usepackage{amsmath}
\usepackage{amsfonts}
\makeatletter
\renewcommand{\@biblabel}[1]{\quad#1.}
\makeatother

\usepackage{lastpage,fancyhdr,graphicx}
\usepackage{epstopdf}
\fancyheadoffset[L]{2.25in}
\fancyfootoffset[L]{2.25in}

\usepackage{color}

\definecolor{Gray}{gray}{.25}
\usepackage{subfig}
\usepackage{graphicx}
\usepackage[acronym]{glossaries}
\usepackage{sidecap}

\usepackage{wrapfig}
\usepackage[pscoord]{eso-pic}
\usepackage[fulladjust]{marginnote}
\reversemarginpar

\begin{document}
\vspace*{0.35in}

% title goes here:
\begin{flushleft}
{\Large
\textbf\newline{Towards Green AI-Native  Networks:\\ Evaluation of Neural Circuit Policy for \\ Estimating Energy Consumption of Base Stations}
}
\newline
% authors go here:
\\
Selim Ickin\textsuperscript{1},
Shruti Bothe\textsuperscript{1},
Aman Raparia\textsuperscript{2},
Nitin Khanna\textsuperscript{2},
Erik Sanders\textsuperscript{2}
\\
\bigskip
\bf{1} Ericsson Research
\\
\bf{2} Ericsson AB
\\
\bigskip
\{selim.ickin, shruti.bothe, aman, nitin.b.khanna, erik.sanders\}@ericsson.com

\end{flushleft}

\newacronym{RAN}{RAN}{Radio Access Network}
\newacronym{NCPs}{NCPs}{Neural Circuit Policies}
\newacronym{LTCs}{LTCs}{Liquid Time-Constant Networks}
\newacronym{LSTM}{LSTM}{Long Short Term Memory}
\newacronym{ML}{ML}{Machine Learning}
\newacronym{KPIs}{KPIs}{Key Performance Indicators}
\newacronym{MSE}{MSE}{Mean Squared Error}
\newacronym{MLOps}{MLOps}{Machine Learning Operations}
\newacronym{OAM}{OAM}{Operations, Administration and Maintenance}

\section*{Abstract}
Optimization of radio hardware and AI-based network management software yield significant energy savings in radio access networks. The execution of underlying \acrfull{ML} models, which enable energy savings through   recommended actions, may require additional compute and energy, highlighting the opportunity to explore and adopt accurate and energy-efficient ML technologies. This work evaluates the novel use of sparsely structured Neural Circuit Policies (NCPs) in a use case to estimate the energy consumption of base stations. Sparsity in ML models yields reduced memory, computation 
and energy demand, hence facilitating a low-cost and scalable solution. We also evaluate the generalization capability of NCPs in comparison to traditional and widely used ML models such as \acrfull{LSTM}, via quantifying their sensitivity to varying model hyper-parameters\,(HPs). \acrshort{NCPs} demonstrated a clear reduction in computational overhead and energy consumption. Moreover, results indicated that the \acrshort{NCPs} are robust to varying HPs such as number of epochs and neurons in each layer, making them a suitable option to ease model management and to reduce energy consumption in \acrfull{MLOps} in telecommunications. \\

{\textit{\textbf {Index terms}}} - Telecommunication network management, Energy efficiency, Machine learning, Sustainable computing

% now start line numbers
%\linenumbers

% the * after section prevents numbering
\section*{Introduction}
Radio Access Networks (RAN) accounts for about 75\% of the total consumed energy in mobile networks. As the demand for data and connectivity continues to increase, the energy efficiency of base stations becomes a critical concern\,\cite{b1}. In the pursuit of sustainable and environmentally friendly mobile networks, the need to address energy consumption at base stations becomes essential. This is also referred to as green networking\,\cite{9844020}.  Traditionally, significant efforts to improve energy efficiency were  related to optimization of energy per bit via hardware and software enhancements in the network elements. In the evolution towards an AI-Native network architecture, the significance of \acrfull{ML} models has heightened as they play an important role in enabling advanced functionalities, intelligent decision-making, and adaptive capabilities within telecommunication networks. The integration of \acrshort{ML} into the future network architecture is essential for realizing the full potential of next-generation telecommunications. AI and ML algorithms are expected to assist functionality of future network elements and improve automated operation and decision-making capabilities\,\cite{ainative}. The increasing demand for reduced carbon footprint and energy consumption in mobile networks motivates the search for novel and eco-friendly strategies for the underlying enabling ML technologies\,\cite{sustainable-ai}. Improving the efficiency of ML models deployed in base stations is an opportunity to reduce the carbon footprint associated with the expanding infrastructure for future mobile networks. The challenge is that computation and storage requirements, as well as model maintenance cost increase with the number of ML models amplifying the complexities in managing \acrfull{MLOps} and making the path to achieving energy efficiency in base stations. The massive number of ML models, each tailored to handle diverse tasks and adapt to varying data distributions, introduces a significant obstacle. Statistical properties of the input data change over time for instance due to base station (re)configuration as a result of dynamic optimization or due to the changing user profiles. To maintain optimal performance, ML models needs to learn from datasets in an efficient way, and to perform continuous adaptation to distribution shifts. Adapting the ML models to changing data distributions manually necessitate continuous hyper-parameter tuning. This is known to be a costly process, in terms of time and energy consumption as a result of potential training iterations during model (re)building phase. \par
Traditional fully connected neural networks result in notably high energy footprints due to the increased computation required for matrix multiplication with every connection between neurons situated in consecutive layers. In response to the escalating challenges of MLOps, including the need for constant adaptation and fine-tuning of models, nature-inspired sparsely structured neural network architectures are starting to be explored\,\cite{nature-inspired-algs}. The role of sparse neural architectures for efficient model training was first introduced in a seminal paper called Optimal Brain Damage\,\cite{optimalbraindamage}, where neuron edges were identified and noncontributing model parameters were iteratively pruned to reduce the computation overhead. Recently, as a result of closer look into the nervous system of a microscopic nematode, C. elegans\,\cite{c-elegans}, \acrfull{LTCs} and \acrfull{NCPs} have emerged as a promising solution. This architecture not only enhances the generalization capabilities of ML models but also presents a novel approach to mitigating the frequent need for model retraining. By dynamically adapting key model components during inference, \acrshort{NCPs} offer a potential breakthrough in minimizing the computational and human efforts required for maintaining an ever-expanding array of ML models. Moreover, continuous time networks\cite{ctnn} have shown to provide bounded estimations on especially irregularly sampled, continuous-time, timeseries datasets, with improved robustness that make them highly suitable for telecommunication use cases. They can adapt to varying time intervals between data points, which is useful for tasks in the telecommunication domain where data collection is not strictly periodic. \par  

This paper captures NCPs application in the telecom domain to primarily propose neural architectures that are small, sparse, computationally and energy efficient without compromising on the accuracy. The benefits may not show comparable improvements in the accuracy domain, but have a significant improvement when it comes to model size, number of parameters to be trained, energy consumed as well as the need for manual effort to define the hyper-parameters\,(HPs) of an ML model. The main motivation of this research is also to explore and provide insights to ML models that  help meeting sustainability goals in AI-native future network design. 

As we go deeper into a detailed evaluation of \acrshort{NCPs} within the context of an \textit{energy estimation} use case in telecommunication context, we aim to understand the practical implications and benefits of this innovative neural architecture in addressing the complex relation between accuracy (in the context of stability against variations in the model HPs and the data distribution), memory allocation, compute efficiency, and overall energy consumption related to resource utilization of these models. Therefore, the summary of the contribution is as follows: i) we evaluate NCPs, via an empirical study, in the context of real mobile network use case with the focus on energy consumption, accuracy, and hyper-parameter tuning efforts of \acrshort{ML}; ii) we present that estimating energy consumption from solely network performance measurement counters is possible, which reduces the data collection, pre-processing, and storage requirements of energy measurement data collected from radio units; iii) we evaluate the impact of noise and data drift on the model accuracy.  \par 
The paper is structured as follows: Section II presents the state-of-the-art on \acrshort{LTCs} based \acrshort{NCPs}, and describes in what aspects they stand out as compared to traditional ones such as \acrfull{LSTM}. Section III presents the related work on ML models that are suitable for sequential problems, such as LSTMs, \acrshort{LTCs} and \acrshort{NCPs}. We also list a few relevant papers on time series based telecommunication use cases. In Section IV, research methodology and evaluation metrics are presented. Section V presents the results, and Section VI provides concluding remarks and suggests potential directions for future research.  
\section{State of the art}
Linear regression models have extrapolation capabilities and are known to be powerful in modeling rather simple problems. For more complex problems  with high dimensional input data and non-linear relationship between input and output, ensemble tree- and Neural Networks-based models are preferred. Ensemble based tree model family that involves Random Forest\cite{randomforest} and XGBoost\cite{xgboost} are good and robust  estimators, however they are known to have challenges in timeseries formulation as they do not have memory capabilities. It necessitates preparing the dataset samples in a way that every data sample is flattened and contains all features in fixed length sliding windows to overcome the memory limitations. Therefore, additional data pre-processing step needs to be performed, which itself is costly and energy consuming. NN models alone do not have memory capabilities either, however they can serve as building blocks for Recurrent Neural Networks (RNNs), and its successors such as LSTMs\cite{lstm}. \par 
\subsection{Continuous Time Recurrent Neural Networks (CT-RNN)}
RNNs can operate on discrete time or on continuous time \cite{b1.5} depending on its architecture. In real-time systems, data is collected continuously, and events from a base station or IoT sensor can be recorded at any moment, unlike fixed time intervals. This often results in irregularly sampled data due to factors like sensor type, sampling frequency, cost, or sensor aging. Instead of using traditional RNNs that operate in discrete time, continuous time neural networks are recommended to better handle data received at any time and to learn temporal dynamics. Training these continuous time neural networks can be challenging since they require differential equation solvers and backward propagation over time. However, they can be more efficient due to the time constant, which helps streamline the training process. In the Ordinary Differential Equation (ODE) of CT-RNN \cite{b1.5}, the time constant parameter $\tau$ provides a more stable neural network as it bounds the activations, thus the outcome of the model, to prevent extreme values. These models are often trained with a batch size of one. Therefore, structured batch input or batch size selection are not needed.

\subsection{Liquid Time Constant Neural Networks (LTC)}
LTC, first introduced by MIT in 2020 \cite{b2} is a new class of CT-RNN. LTC’s have shown outstanding performance over other well-known techniques in navigation of unknown territory with autonomous driving\,\cite{autonomousdriving}. LTCs are trained using an ODE solver followed by Backward Propagation Through Time (BPTT). \acrshort{LTCs} adapt without needing retraining, therefore is expected to greatly reduce the computation overhead. \par

LTC networks are comprised of first-order dynamical systems with interconnected nonlinear gates. In LTC, the neural network is multiplied by the difference of $A$, a biasing parameter and the hidden state of the neural network, i.e., $(A-x(t))$ as shown in Eq.\,\ref{Eq2} to \ref{Eq4}. This way, the state of the neural network becomes a function of hidden state. In addition, in \acrshort{LTCs}, the coefficient of $x(t)$ is defined by a neural network in contrast to CT-RNN, where coefficient was only a constant, $\tau$.

\begin{equation}
\frac{dx(t)}{dt} = - \frac{x(t)}{\tau} + S(t) \quad S(t) \in \mathbb{R}^M
\label{Eq2}
\end{equation}

Eq.\,\ref{Eq2} describes the change in the hidden state  $x(t)$ over time, with a time constant
$\tau$ and an input signal, $S(t)$.
\begin{gather}
S(t) = (f(x(t), I(t), t, \theta))(A-x(t))
\label{Eq3}
\end{gather}

Eq.\,\ref{Eq3} defines the input signal $S(t)$ as a function of the hidden state $x(t)$, input 
$I(t)$, time $t$, and parameters 
$\theta$, modulated by the difference between A and $x(t)$. Eq.\ref{Eq4}  shows how the hidden state $x(t)$ evolves with a variable coefficient. This coefficient is not constant anymore but dependent on $x(t)$, $I(t)$, time $t$, and NN model parameters, $\theta$ as given in  Eq.\ref{Eq5}.
\begin{equation}
\begin{split}
\frac{dx(t)}{dt} = - [ \frac{1}{\tau} + (f(x(t), I(t), t, \theta))] \ x(t) +\\ f(x(t), I(t), t, \theta)A 
\end{split}
\label{Eq4}
\end{equation}
\begin{gather}
\tau_{\text{sys}} = \frac{\tau}{1 + \tau (f(x(t), I(t), t, \theta))}
\label{Eq5}
\end{gather}
This property helps adaptation, which is a variable time constant which itself is learned via neural network. This way, even if the inputs to the system goes to extremely high or low abnormal values, the time constant \( \tau_{\text{sys}} \) of the neural network system becomes stable, bounded and adaptive to the input and state. Moreover, the state of the neurons becomes bounded as well. \par 

\begin{figure*}[h!]
\includegraphics[width=\linewidth]{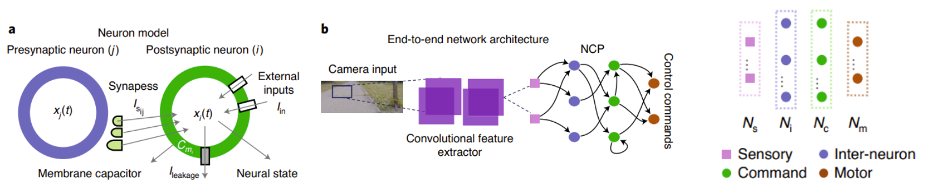}
\caption{NCP architecture and functionality. \cite{b2}}
\label{fig:NCPArch}
\end{figure*}

In summary, the key difference between \acrfull{LTCs} and CT-RNN’s is that \acrshort{LTCs} allows the parameters of the equations to change over time, with an alteration of the responsiveness of the neurons. The advantages of \acrshort{LTCs} can thus be summarized as follows: 
\begin{enumerate}
    \item \acrshort{LTCs} have input-dependent varying time constant. This liquidity property of \acrshort{LTCs} helps to adapt a pre-trained ML model to unseen datasets even after training, without necessitating retraining or fine-tuning. This is realized with a negative feedback mechanism within the synapse of two neurons (pre- and post-synaptic neuron). This makes them distinguish against their preceding CT-RNN models.
    \item Due to its adaptive time constant nature, \acrfull{LTCs} keep track of the hidden state of the network, and provide bounded results without diverging, which makes them prone to over-fitting. 
    \item Massive sparsity of \acrshort{LTCs} enables smaller models (significantly fewer numbers of model parameters), making them storage,  memory, and computation-friendly. 
\end{enumerate}

\subsection{Neural Circuit Policy (NCP)}
\acrshort{LTCs} are the building blocks of \acrshort{NCPs}, an architecture designed with the inspiration of the  microscopic nematode, C. elegans. It has very few neurons in its nervous system and can generate unexpectedly complex dynamics. NCP’s have 90\% sparsity in the intermediate neurons with unique connection topology between inter-neurons and command neurons, and has therefore less storage and memory requirements. LTC architecture is given in  Fig.\,\ref{fig:NCPArch}\,\cite{b2}. 
The neural dynamics of NCPs are also expressed through continuous-time ODEs. Fig.\,\ref{fig:NCPArch}(a) illustrates the neural state \( x_i(t) \) of a postsynaptic LTC neuron \( i \), which receives input currents from a presynaptic neuron \( j \). The neural state is influenced by the sum of various inflows and outflows within the cell. The external input currents are denoted as \( I_{\text{in}} \), while \( I_{\text{leakage}} \) represents the leakage current. Synaptic currents \( I_{s_{ij}} \) are determined by an input-dependent non-linearity \( f \), which is a function of the presynaptic neural state \( x_j(t) \) and its synaptic parameters.

Fig.\,\ref{fig:NCPArch}(b) depicts an end-to-end NCP network on an image dataset in \cite{b2}. The network processes camera inputs through a series of convolution layers, transforming them into a latent representation. This latent representation is then utilized by the designed NCP to generate control actions and issue control commands.
NCP is a network of LTC systems consisting of $4$ layers of specific architecture where intermediate neurons are highly sparse and recurrent. 

\subsection{Architectural differences in LSTM and NCPs}

The architectural differences between LSTM networks and NCPs lie primarily in their internal structures and learning mechanisms. LSTM networks are a type of RNN designed to capture long-term dependencies in sequential data. They consist of memory cells and gates that regulate the flow of information, allowing them to retain and utilize information over extended time periods. LSTMs excel at tasks that require memory retention, such as language modeling and predictive tasks on timeseries datasets. However, they can be computationally intensive and struggle with handling data drift.\par 
On the other hand, NCPs are inspired by liquid time constant neural networks and exhibit a sparse, interconnected architecture. They feature first-order dynamical systems with non-linear gates, enabling efficient computation and adaptability. NCPs leverage continuous-time dynamics and unique connection topologies to enhance model generalization and handle data drift more effectively, especially in scenarios with limited training data. Additionally, NCPs offer inherent adaptability during inference without requiring extensive retraining.\par
%, making them well-suited for dynamic environments like telecommunications networks
In summary, while LSTM networks excel at capturing long-term dependencies and memory retention through memory cells and gates, NCPs leverage sparse, continuous-time dynamics to achieve efficient computation, making them well-suited for dynamic environments such as mobile communication networks. 
\section{Related work}
The paper ``Advancements in Time Series Forecasting" \cite{b3} explores how algorithmic limitations have hindered Time Series (TS) forecasting, despite recent strides in Deep Learning (DL). Through a intensive literature review, the authors highlight the challenges and applications of TS forecasting algorithms. They focus on Neural Ordinary Differential Equations (NODEs) and LTC networks, which offer enhanced adaptability. These algorithms show promise in addressing TS forecasting shortcomings compared to DL. This paper also supports an argument where NCPs inspired by ODEs cannot model randomness and instantaneous changes. However, Stochastic Differential Equations\,(SDEs) \cite{sdes} are more flexible as they handle randomness. \par 
There is previous work where LTCs are used in the context of telecommunication use cases. In \cite{b4}, the authors propose utilizing an LTC network to forecast blockage status in a millimeter wave (mmWave) link, using only received signal power as input. The authors explore the similar LTC network and compare it to DNNs, Support Vector Machine (SVM) and Multinomial Logistic Regression (MLR). The prediction accuracy increases significantly (by $12$\%, $39$\%) when predicting even just slightly further into the future ($t$+5, $t$+10) indicating that LTCs are a fit for handling TS (forecasting) problems. Authors in \cite{b5}, explore the NCP architecture for improving time-series prediction. Traditional models incur high training costs due to their size, prompting the development of a novel machine learning system. This system predicts cellular traffic using only $225$ parameters, a significant reduction compared to standard models. Results indicate that a $64$\% sparse model outperforms current ones (LSTM) threefold, offering a precise performance boost. Moreover, the framework provides automatic parameter optimization for optimal sparse model selection and employs a distributed setup to share model details among cells with similar behaviors, thereby reducing training time and costs. This paper supports the conclusion that LTC model had significantly less computation overhead. \par 
\par 
These additional insights highlight the unique contributions of our study in leveraging LTCs and NCPs for energy-efficient modeling in telecommunications networks. While LSTM models are commonly used in prediction tasks, our study explores the use of NCPs for improved energy efficiency in modeling. By analyzing model performance,  computational overhead, and energy consumption we aim to demonstrate the benefits of using NCPs also in estimating energy consumption of a base station. In this paper, we demonstrate and quantify the level of energy savings that can be achieved by replacing LSTMs with NCPs in energy estimation of a base station. \par 
%TimeGPTs....
\begin{figure*}
    \centering
    \includegraphics[width=0.9\linewidth]{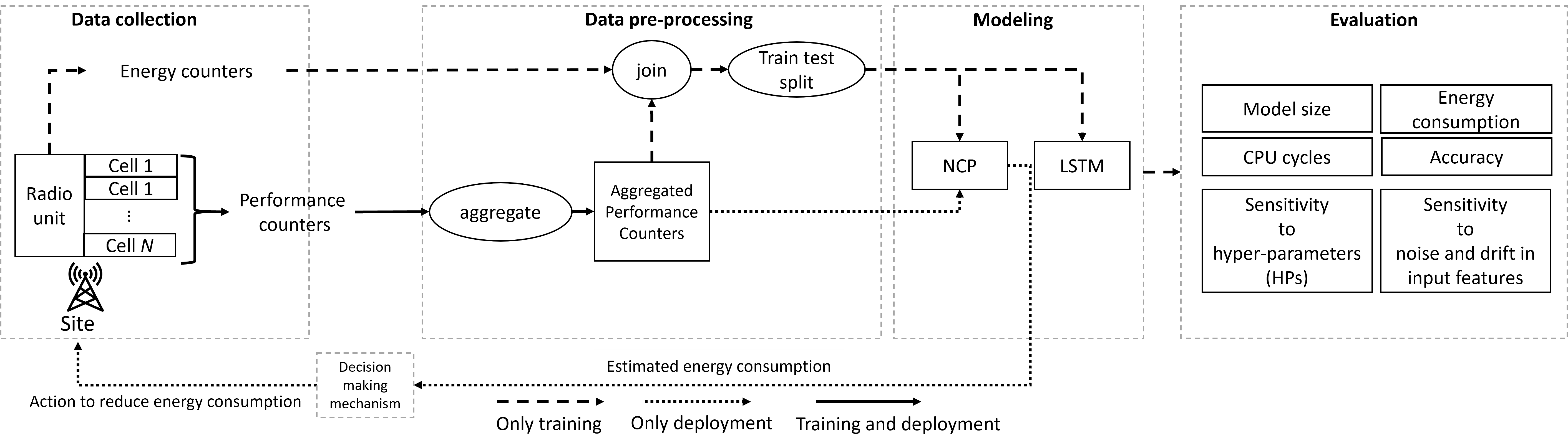}
    \caption{Flow chart illustrating the e2e pipeline of the ML models in training and inference phases. It also illustrates each process is potentially located in the mobile network. }
    \label{fig:method-chart}
\end{figure*}
\section{Methodology}
We selected a use case that aims to estimate the energy consumption of a base station from network performance counters observed in the same base station. The methodology of the study is illustrated in a flow chart in Fig.\,\ref{fig:method-chart}. It consists of data collection, pre-processing, modeling and evaluation steps. In the training phase, both energy counters and performance measurement counters are used since the goal is to train a model that estimates the energy consumption from the input performance features. In the deployment phase, only performance measurements are collected and then energy is estimated from the performance counters. The reason for this is that cell energy consumption measurements are technically harder to obtain than the performance management counters, due to their size and location of the measurement points, therefore an energy estimation capability solely from performance management counters would reduce MLOps requirements such as data collection and pre-processing time. We train LSTM and NCP models separately, and then compare their performance in the evaluation phase. The performance and energy data is typically collected at the RAN and then sent to a data pre-processing unit. The pre-processing unit can then send the data to a data analytics function \cite{rel18}, where further operations on the dataset, such as analytics and ML modeling can be hosted. The result of the ML model, \textit{i.e.}, estimated energy consumption, can then be sent to the network analytics consumer, so that right decisions and actions (e.g, reconfiguration of the base station) can be taken to reduce energy consumption.   

\subsection{Data collection}
The data is collected from various sources such as LTE and 5G New Radio (NR), and is a multi-variate timeseries dataset. The dataset comprises of performance counters for all the required KPIs and cell/site information related to utilization of the physical resource blocks,   data volume, number of connected users and active sessions, transferred data volume, throughput, and signaling overhead. The dataset \cite{data} had measurements from both uplink and downlink, i.e., from users to the base station and vice-versa, respectively. These performance counters serve as the basis for calculating KPIs at the cell level. To integrate the energy data with the KPI data, aggregation of cell data at a higher level is necessary. This aggregation can be achieved through mappings between cells and radio units, which can be derived from parameters. The energy consumption is measured at a physical hardware unit. Mapping logical cell level data to physical hardware level data necessitates additional data pre-processing. A strong energy estimator model that uses solely performance measurement counters is expected to reduce the necessity of such additional data pre-processing. 
\begin{figure}
    \centering
    \includegraphics[width=\linewidth]{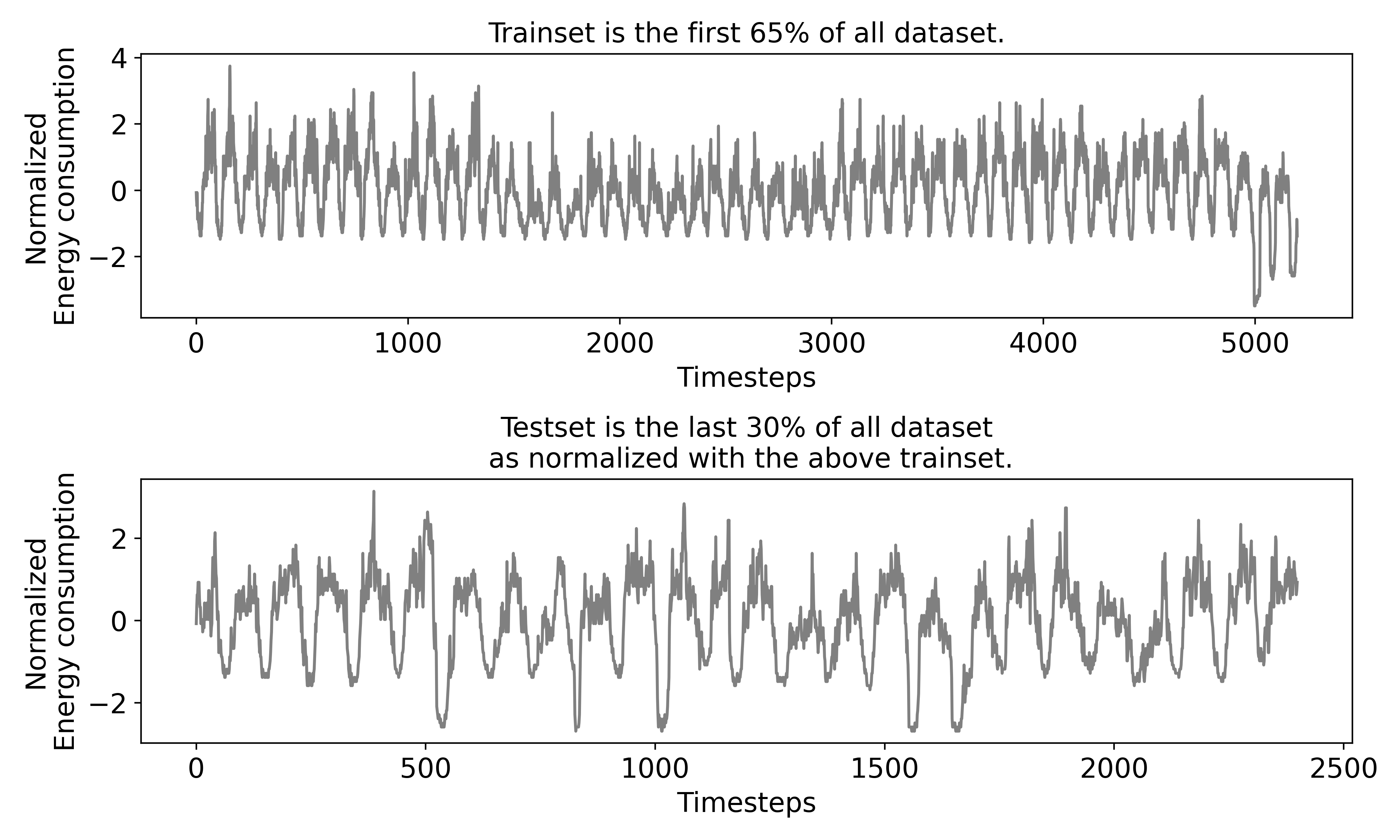}
    \caption{Snapshot of normalized energy consumption of a base station is given in the figure (top: trainingset, bottom: testset), where there exists data drift between the training and testset.}
    \label{fig:data_sample}
\end{figure}
\subsection{Data pre-processing}
Data pre-processing involved handling information collected every $15$ minutes from $249$ base station sites. If there were multiple cells in a single radio unit, we aggregated the performance counter values for those sharing the same unit and \emph{joined} them with the energy consumption data. The pre-processing resulted in one set of measurement data for each site, each radio unit, and each $15$-minute time interval. We then combined all the data related to each site into a file, creating a total of $249$ files. Missing values are replaced by the preceding timestamp value.\par 
This dataset comprised data from a subset of sites, specifically chosen from a city center along with its $50$ neighboring sites within approximately five kilometers. It encompassed multivariate time series data of various KPIs and energy consumption values collected from different types of radio units. To maintain consistency in energy consumption and KPI patterns, we filtered the dataset to include only one type of radio unit, which had the highest number of data points. However, some sites had longer data intervals compared to others. Consequently, we selected a time interval common to all sites. Each site and radio unit type contributed $8000$ data points. Despite filtering the dataset, we observed various data patterns related to variations in data distributions, ranging from light to heavy data drift. To address this variability, we applied K-means clustering, resulting in the identification of five distinct clusters. The selection of the number of clusters was determined based on the coherence score. Subsequently, five clusters, characterized by differing levels of data drift, and we exploited the sites that are in the cluster of small data drift. ODEs, which the NCP models operate on, are known to be not able to cope with sudden changes in timeseries, and the data drift on other sites contained such characteristics. Since coping with extreme data drift needs additional orthogonal ML methods in the area of continual learning and domain adaptation, we leave dealing with the heavy drift scenarios as part of future work. \par 
The training set size was the first 65\% of the dataset. We reserved the last 30\% of the dataset as the test set. There were $5200$ samples in the training set and $2400$ samples in the testset. An example dataset is visualized in Fig.\,\ref{fig:data_sample}.  The mean values of the  trainset were $0$ since we used the scikit-learn Python library and   \textit{StandardScaler} function to transform the data to standard normal distributed data, and the mean values of testset was $-0.008$. This indicates that the values in the test set was slightly lower than the training set. 

\subsection{Modeling}
The NCP architecture used in this paper had a sparsity of $0.9$, i.e., 90\% of the model parameters were pruned by model architecture with NCP wiring between neurons. The learning rate of LSTM and NCP models were both set to $0.005$. We perform training on both models with various HPs such as number of epochs=\{$50$, $100$, $200$, $400$\}, number of neurons at the intermediate layer {$16$, $32$, $64$, $96$\}, total layer count of $4$. The latter two defined the total number of model parameters, i.e., number of weights in the model. In order to test the behaviour of the models in an over-training scenario, we also did an experiment by setting the epoch count to $800$, and number of neurons in the intermediate layers to $16$.

\subsection{Evaluation} We compared two different NN based ML models that are suitable for timeseries data: LSTM and NCPs. The comparison is obtained with respect to their model accuracy on the same test set data, computation overhead, as well as energy consumption during training. We used R2-score and \acrfull{MSE} for evaluating the accuracy of the models as it was formulated as a regression problem. R2-score ranges between $0$ and $1$, where $1$ is the highest accuracy, where $0$ is the worst. \acrshort{MSE} is the mean of the squares of the difference between estimations and actual values, hence small value for the metric is preferred. We present the \acrshort{MSE} values via tail analysis, since the rare high values in energy consumption are equally  interesting as the average values. Therefore, it is expected from an ML model to accurately estimate the instances with high energy consumption values to enable preventive actions in advance. Therefore, we in particular quantify the \acrshort{MSE} in the values above $90^{th}$ percentile. \par 

As one final test for the aim of quantifying model robustness of the ML models to noise and drift, and we applied various levels of Gaussian noise to the input features and drift to the label, i.e., energy measurements in the testset. The motivation for adding noise is to observe how a model that is trained on the original training set would perform on a noisy testset. We obtained the noisy data by adding to the original test data a Gaussian noise with mean$=0$ and variance$=(max(D_{\textrm{testset}})-min(D_{\textrm{testset}})) * \epsilon$, where $D_{testset}$ are all data samples of testset, and $\epsilon=\{0.025, 0.05, 0.1\}$. The motivation for adding drift is to observe the impact of a scenario when a hardware unit is upgraded with a more energy-efficient one consuming statistically less energy. We obtained the drifted data by changing the values of the original test data by subtracting ($max(D_\textrm{testset})-min(D_{\textrm{testset}})) * \epsilon$, where $D_{testset}$ are all data samples of testset, and $\epsilon=\{0.01, 0.05, 0.075\}$. After obtaining the datasets, the distribution differences between the original training set and the obtained testsets are quantified via Kolgomorov Smirnov (KS) test statistics using the \emph{$ks\_2samp$} function in Python\,\cite{KS}, \cite{ks-python}.  \par 
We then compared the two models' performances. Such testing against robustness was previously used in \cite{addnoisetotestset}. Together with the number of model parameters, i.e., neuron weights of ML models, the energy consumption related to the resource utilization during the training of all models were also measured via Kepler tool \cite{b6}, \cite{b7}. Experiments were performed in Kubernetes with a maximum resource allocation of $40$Gi and $10$ CPUs.

\begin{figure*}%
    \centering
    \subfloat[Overall model performance (R2-score). Cyan: 50 epochs, red: 100 epochs, magenta: 200 epochs, purple: 400 epochs, gray: 800 epochs.\label{overallacc}]{
        \includegraphics[width=0.47\linewidth]{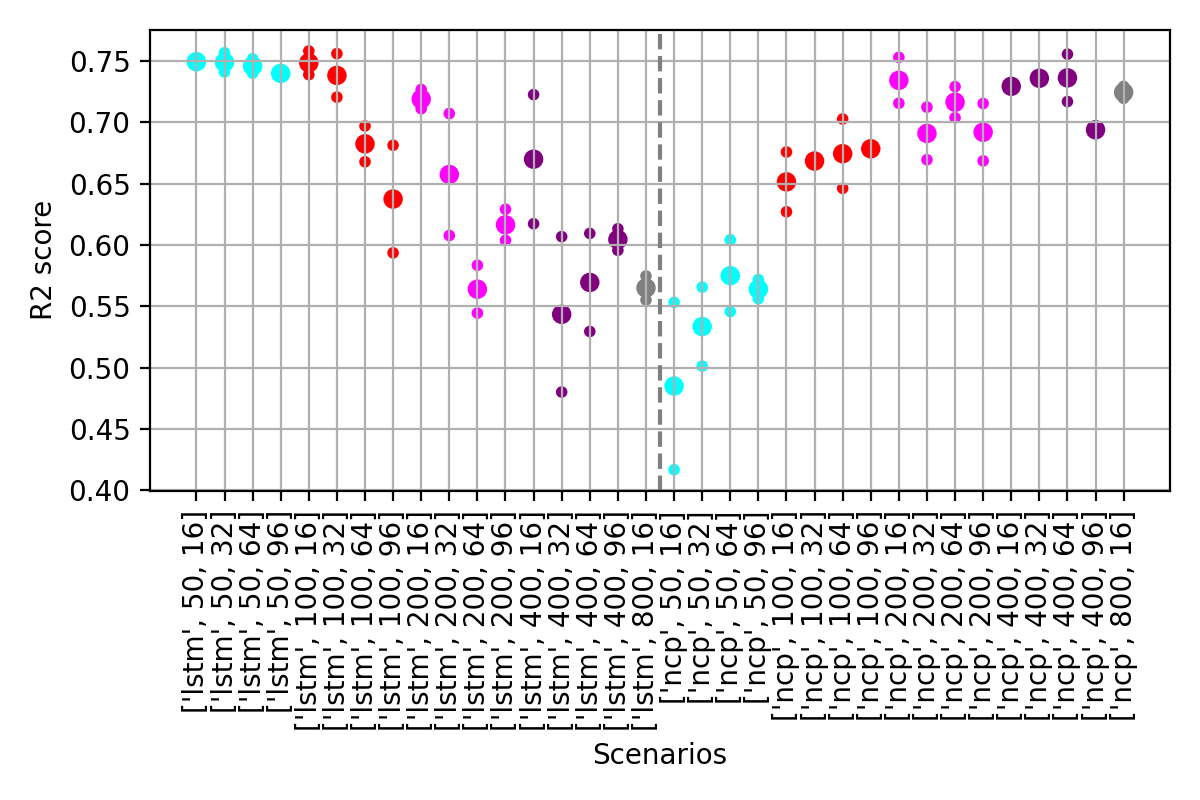}
    }
    \hfill
    \subfloat[\acrfull{MSE} on the $90^{th}$ percentile. Cyan: 50 epochs, red: 100 epochs, magenta: 200 epochs, purple: 400 epochs, gray: 800 epochs.\label{tailacc}]{
        \includegraphics[width=0.47\linewidth]{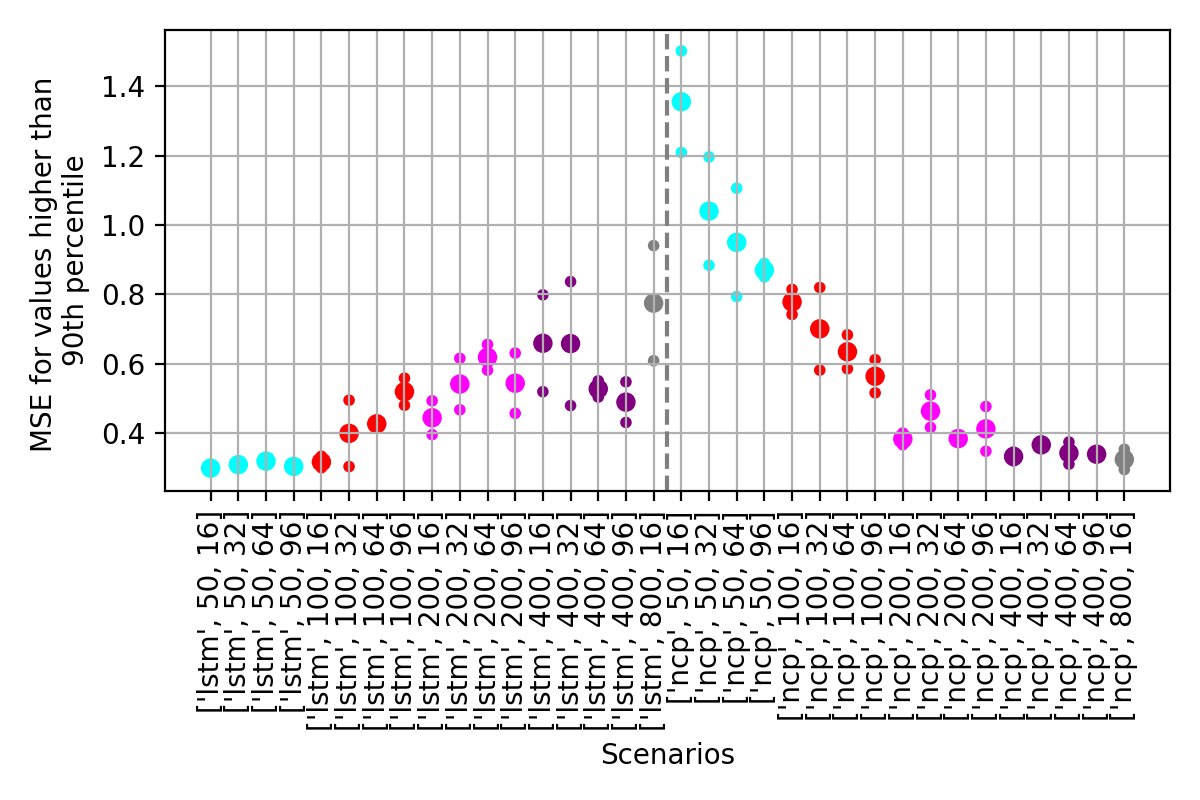}
    }
    \caption{Comparison of model performances using R2-score and MSE in different experiment settings.}
    \label{fig:acc}
\end{figure*}

\section{Results}
The results obtained from the scenarios are illustrated via upper and lower confidence bounds (95th percentile) in Fig.\ref{fig:acc}. X-axis indicates the model type, number of epochs and the number neurons in the intermediate layer. A  large circle indicates the mean, and smaller circles above and below the large circles are the corresponding confidence bounds indicating the deviation around the mean for each scenario. Each circle group of the same color represents group of scenarios belonging to the experiments conducted with the same number of epochs. Fig.\,\ref{fig:acc}(a) and Fig.\,\ref{fig:acc}(b) illustrate the overall model performance and the performance of the model at the samples above $90^{th}$ percentile, respectively. The LSTM performance heavily depends on the choice of hyper-parameters\,(HPs) showing evidences that it overfits as the number of model parameters and epochs increase. NCPs manifest more robust results with varying HPs. Moreover, interestingly, NCPs yielded more confident and accurate results with the increase in the epochs and number of neurons. One conclusion from this observation is that assigning high number of epochs and number of neurons at the intermediate layer of NCP provides high accuracy guarantees as well as lower computation overhead, model parameter size, and energy consumption. While in LSTM, the accuracy is highly sensitive to the choice of such hyper-parameters hence causes additional tests to choose the best ones. Moreover, each experiment with LSTM costs higher in terms of computation and consumed energy required for training those experimental models.  \par 
The best performing NCP model that had $64$ neurons in the intermediate layers, and when trained $400$ epochs reached a mean highest R2-score of $0.736$. Although this scenario had also low \acrshort{MSE} on the values above the $90^{th}$ percentile, the scenario that had $16$ neurons in the intermediate layer trained with $400$ epochs had the least \acrshort{MSE} with $0.33$ (vs $0.34$ achieved with $400$ epochs and $64$ neurons). Another good performing configuration for NCP had $32$ neurons and $400$ epochs. This yielded also an overall R2-score of $0.736$ and but a tail performance of $0.36$. The NCP variant with intermediate layer neuron size of $16$ and epoch count $200$ also yielded a good R2 score, $0.734$. On the other hand, the best performing LSTM model that had $100$ epochs and $16$ neurons at the intermediate layers yielded an R2-score of $0.748$ with a $90^{th}$ percentile tail performance measured via \acrshort{MSE} of $0.31$. LSTM model with $16$ neurons at the intermediate layer with $50$ epochs reached a similar performance, too. In general, the R2 score and \acrshort{MSE} differences between the best case models were not considered significant, still LSTM performed marginally better in overall, given the settings of the experiment. Over-training of LSTM with $800$ epochs, as depicted with gray color, caused over-fitting, yielding an R2-score of $0.56$, while training of NCP with the same number of epochs ($800$) sustained a high R2-score of $0.724$. We see a similar performance on the tail analysis where LSTM and NCP yielded \acrshort{MSE} of $0.77$ and $0.32$, respectively.  \par 

\begin{figure*}%
    \centering
    \subfloat[Energy consumption of both ML models in different scenarios. Cyan: 50 epochs, red: 100 epochs, magenta: 200 epochs, purple: 400 epochs, gray: 800 epochs.\label{fig:energy}]{
        \includegraphics[width=0.47\linewidth]{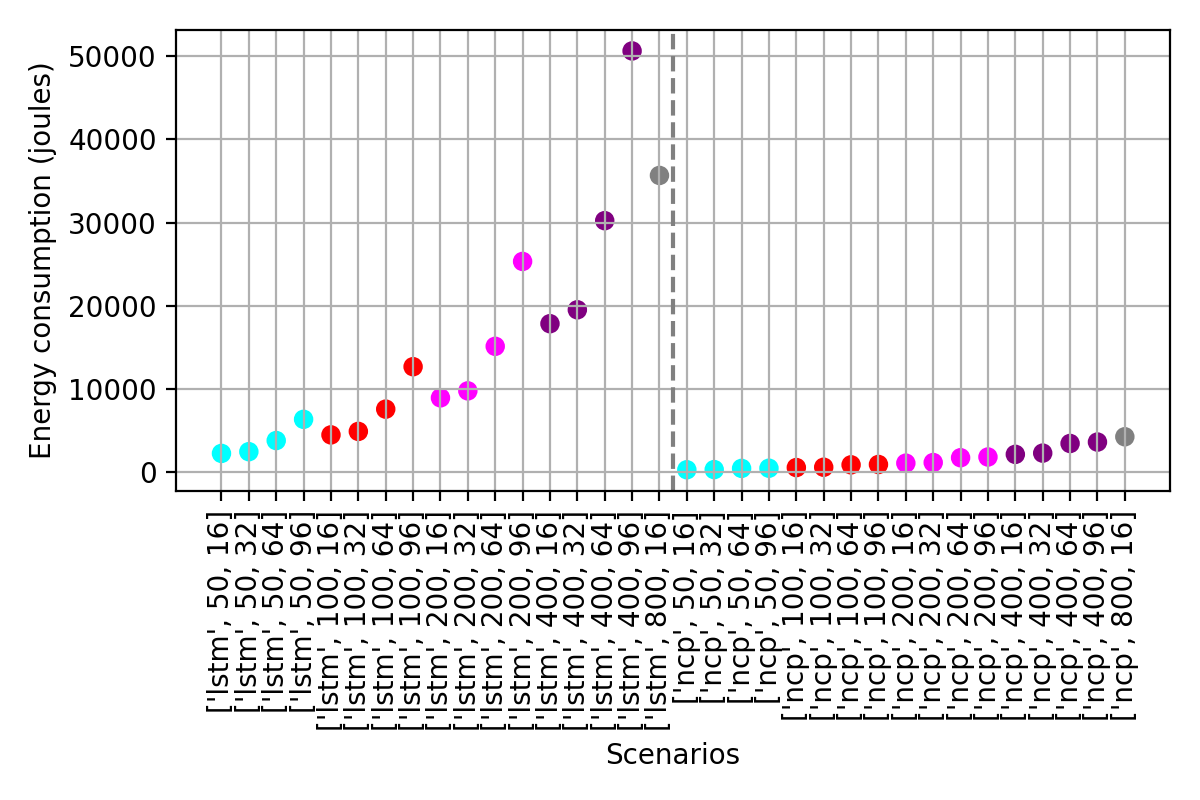}
    }
    \hfill
    \subfloat[Number of model parameters of both models in different scenarios. Cyan: 50 epochs, red: 100 epochs, magenta: 200 epochs, purple: 400 epochs, gray: 800 epochs.\label{fig:neuronsize}]{
        \includegraphics[width=0.47\linewidth]{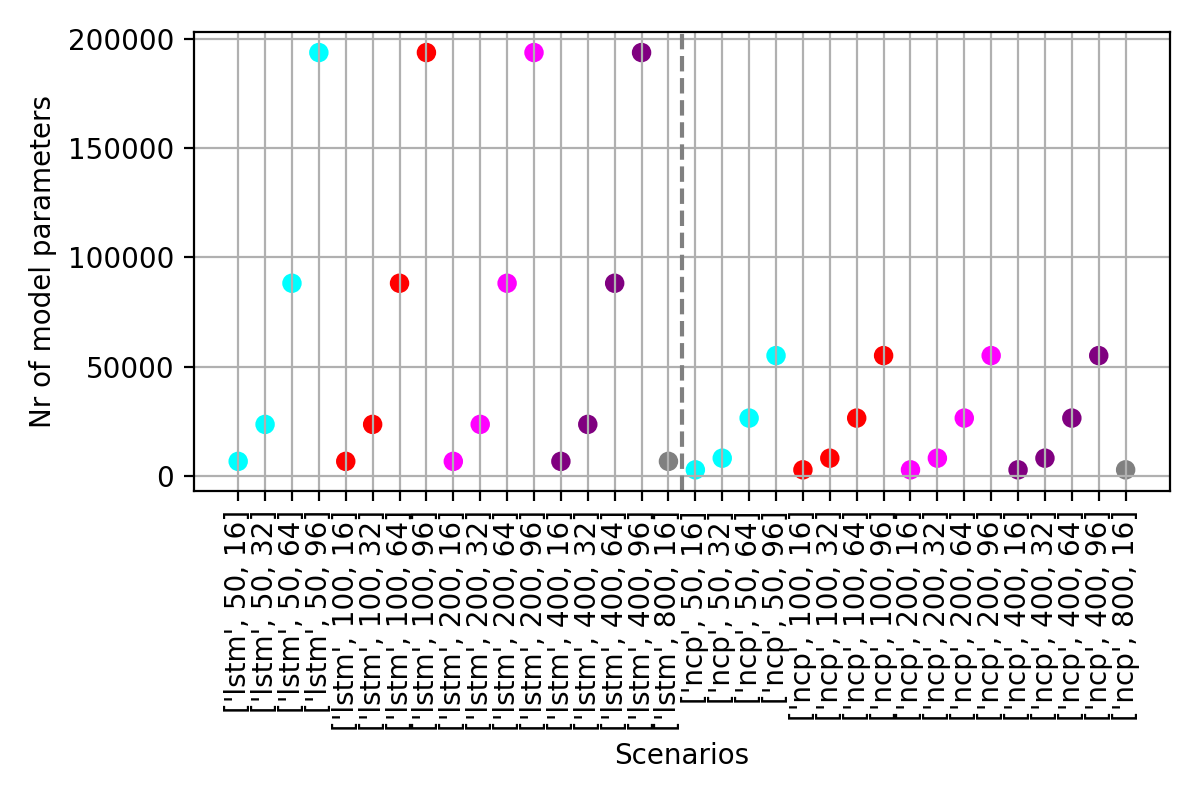}
    }
    \caption{Energy consumption and model parameter size of LSTM and NCP models in different scenarios.}
    \label{fig:overhead}
\end{figure*}

The energy consumption and the model parameter size of LSTM and NCP ML models are given in Fig.\ref{fig:overhead}. Fig.\ref{fig:overhead}(a) presents the scenarios and the corresponding consumed energy values for each scenario. Similar to Fig.\,\ref{fig:acc}, each group of circles of the same color represent the results that belong to the experiments performed with the same number of epochs. The scenario of LSTM with $16$ neurons and $100$ training epochs yielded an energy consumption of $4458$ joules, while the scenario of NCP with $64$ neurons for $400$ epochs yielded an energy consumption of $3423$ joules. NCP variant that had $16$ neurons and $400$ epochs also performed well ($0.729$) and yielded an energy consumption of $2116$ joules; again an NCP variant that had $16$ neurons trained for $200$ epochs had good R2-score ($0.734$) and yielded energy consumption of $1058$ joules. These values are still lower than the energy consumption of the smallest but high performing LSTM model with $16$ neurons at the intermediate layer with $50$ epochs, $2229$ joules. The energy consumption of the LSTM increases significantly with the increase in the number of model parameters and epochs, and is order of magnitude higher in the equivalent configuration scenario of NCP. This makes the choice of model training with NCP appealing since the configuration that would yield the best performance is hard to know in advance, and hence requires a sequence of trial and errors to converge to the best hyper-parameters and thus yielding high energy consumption. While in NCP, it would be reasonable to keep the number of parameters high in the first attempt to guarantee both high performance, reasonable energy consumption as well as reducing the time spent for model hyper-parameters in ML operations. 
Fig.\ref{fig:overhead}(b) presents the number of neurons of different scenarios. Best case NCP scenario had $26470$ model parameters, which is significantly less than LSTM scenario with equivalent configuration that had $89129$ model parameters due to the reason that NCPs have highly sparse structure while LSTM models are fully connected.\par 
\begin{figure*}%
    \centering
    \subfloat[Impact of additional noise on the model performance.\label{noise}]{
        \includegraphics[width=0.47\linewidth]{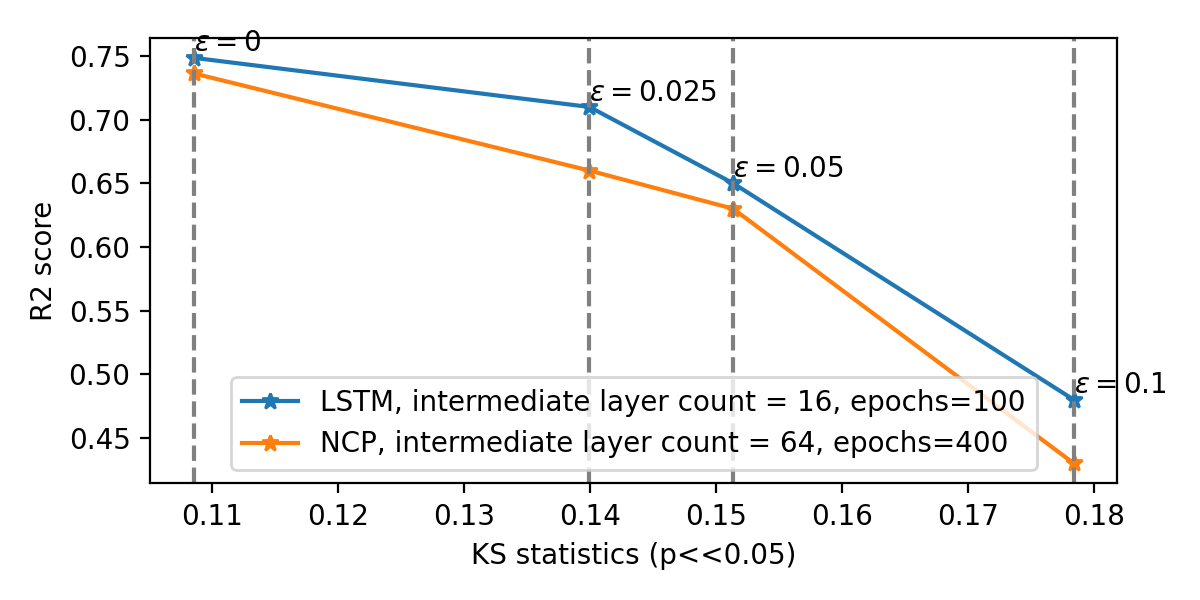}
    }
    \hfill
    \subfloat[Impact of drift on the model performance.\label{drift}]{
        \includegraphics[width=0.47\linewidth]{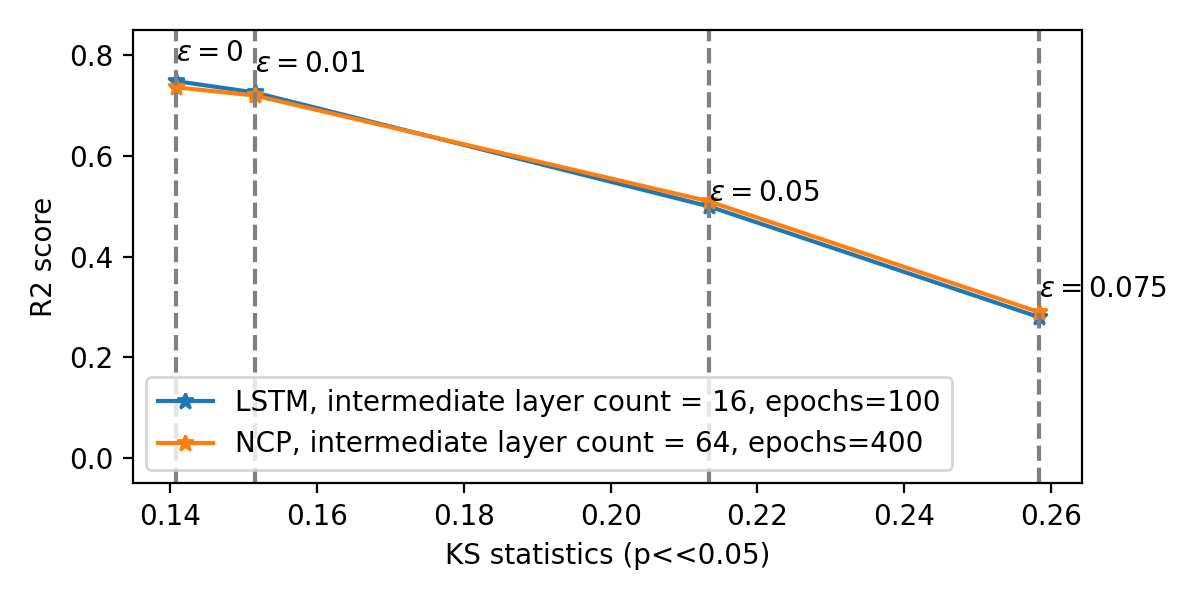}
    }
    \caption{The impact of noise and drift on the model performance (R2-score) is illustrated.}
    \label{fig:sensitivity_to_synthetic}
\end{figure*} 

In order to test the sensitivity of both models with changing distributions in the testset, we performed experiments by adding synthetic noise and synthetic drift. We performed the experiments on the best performing versions of the two models, i.e., LSTM model instance (epochs=100, number of neurons at the intermediate layer = 16), NCP model instance (epochs=400, number of neurons at the intermediate layer = 64). The results from this experiment is given in Fig.\,\ref{fig:sensitivity_to_synthetic}. Blue and orange lines depict the R2-score performance of LSTM and NCP with varying epsilon values, respectively. X-axis is the KS statistics, and a higher KS statistic value corresponds to higher difference between the original testset and the synthetic testset. This quantifies the observed level of drift in the testset. In all cases, $p$ value (at $95^{th}$ percentile) was significantly less than $0.05$. Fig.\,\ref{fig:sensitivity_to_synthetic}(a) illustrates the model's performance with additional noise with varying levels indicated by $\epsilon$. Fig.\,\ref{fig:sensitivity_to_synthetic}(b) illustrates the model performances with varying additional drift with varying levels indicated by $\epsilon$. As can be interpreted from the figures, both models are equally sensitive to the noise and drift in the testset. Hence, we can not conclude that NCPs are robust to additional noise and drift in the testset given the experimental settings and the dataset within the context of this paper.

A high-level evaluation summary is given in Table\,\ref{tab:evaluation_summary}. NCP models are 90\,\% sparse by architecture, and has significantly less number of parameters to train. The NCP model exhibits notably reduced energy consumption related to the reduced resource utilization in comparison to LSTMs, possibly owing to architectural disparities such as neural connection sparsity. This in parallel makes NCPs more energy-efficient than LSTMs. The results indicate that LSTMs and NCPs exhibit comparable performance, with LSTMs showing a slight advantage, particularly with less parameters and shorter training, i.e., small epochs. Both models exhibit good performance in the tail analysis on their best case configurations, ensuring that estimations do not under-shoot for very high energy values, maintaining a higher level of safety. This aspect is crucial for applications in dynamic environments where under-estimated energy consumption is not preferred to minimize missed actions to prevent it. As the model parameter size and number of epochs increase, LSTM models tend to exhibit signs of over-fitting, while NCPs maintain stability in their performance. Selecting the best hyper-parameters for a model in one shot, i.e., without a few trials, is difficult, and hence requires many experiments with different configurations to be performed where each is costly in terms of energy consumption. In that aspect, NCPs are advantageous since setting high number of model parameters and epochs in the first attempt does give performance and energy saving guarantees as compared to trial and error based hyper-parameter tuning in the case of LSTM. Overall, both models showed signs of negative impact on the R2-score with respect to noise and drift on the testset.

\begin{table}[!ht]
\begin{adjustwidth}{-2in}{0in} 
\centering
\caption{{\bf Evaluation summary of models} }
\begin{tabular}{c||c|c|c|c|c|c}
        Model & Model  & Energy &  Best  & Sensitivity to & Sensitivity to the  & Sensitivity to the    \\ 
        name & parameter size  &  consumption & Accuracy & the model & noise in the & drift in the \\ 
        & & & & hyper-parameters & input features &input features\\ \hline\hline 
        LSTM & Higher     &   Higher  &   Marginally higher    &   Higher & On-par & On-par \\ \hline
        NCP  & Lower      &   Lower   &   Marginally Lower     &   Lower  & On-par & On-par \\ \hline

\end{tabular}
\label{tab:evaluation_summary}
\end{adjustwidth}
\end{table}
\section{Conclusion}

In this paper, we evaluate Liquid Time Constant Neural Networks (LTC) and Neural Circuit Policies (NCPs) for estimating energy consumption in telecommunication networks. NCP is recommended as a novel neural network model architecture tailored for energy estimation and potentially for other performance prediction tests in radio access networks. By utilizing real-world data, we showcase the superior performance of NCPs in reducing computation overhead, memory allocation and energy consumption, while sustaining high model accuracy. We showed that NCPs are highly robust to the choice of number of neurons per layer and the epoch count, therefore reducing the excessive energy consumption by bypassing the model hyper-parameter tuning. Instead, a high number of neurons and epochs can be selected and this gives high model accuracy guarantees in contrast to  LSTM. We did not observe additional robustness of NCPs as compared to LSTMs in synthetic noise and drift scenarios which may be investigated further in future work. Our research is driven by the motivation to develop and analyze machine learning models that support sustainability goals in AI-native future network design, emphasizing small, sparse, and energy-efficient architectures. \par We see potential future directions that can be studied on NCPs,  such as understanding the impact of different sparsity levels on the performance, the performance comparison between SDE's and ODE's, as well as exploration of additional novel methods to embed into the NCPs in order to further improve the robustness of the model against noisy and drifted data in the context of telecommunications.  

\section{Acknowledgements}
We would like to thank Konstantinos Vandikas and Daniel Lindström at Ericsson Research for helping us to install Kepler tool in the Kubernetes cluster that enabled energy measurements.

%\nolinenumbers

%This is where your bibliography is generated. Make sure that your .bib file is actually called library.bib
\bibliography{library}

\begin{thebibliography}{10}

\bibitem{b1}
``On the road to breaking the energy curve.'' Ericsson Technical Report, 2022.
\newblock \url{https://www.ericsson.com/4aa14d/assets/local/about-ericsson/sustainability-and-corporate-responsibility/documents/2022/breaking-the-energy-curve-report.pdf}.

\bibitem{9844020}
A.~Clemm and C.~Westphal, ``Challenges and opportunities in green networking,'' in {\em 2022 IEEE 8th International Conference on Network Softwarization (NetSoft)}, pp.~43--48, 2022.

\bibitem{ainative}
``Defining ai native: A key enabler for advanced intelligent telecom networks.'' Ericsson White Paper, 2023.
\newblock \url{https://www.ericsson.com/en/reports-and-papers/white-papers/ai-native}.

\bibitem{sustainable-ai}
K.~E. Yaacoub, O.~Stenhammar, S.~Ickin, and K.~Vandikas, ``Continual learning with siamese neural networks for sustainable network management,'' {\em IEEE Trans. Netw. Serv. Manag.}, vol.~21, pp.~2664--2674, Jun. 2024.

\bibitem{nature-inspired-algs}
H.~C. et~al., ``Nature inspired meta-heuristic algorithms for deep learning: Recent progress and novel perspective,'' in {\em Advances in Computer Vision. CVC 2019. Advances in Intelligent Systems and Computing} (K.~Arai and S.~Kapoor, eds.), vol.~943, Cham: Springer, 2019.

\bibitem{optimalbraindamage}
Y.~LeCun, J.~S. Denker, and S.~A. Solla, ``Optimal brain damage,'' in {\em Advances in Neural Information Processing Systems 2 (NIPS 1989)}, pp.~598--605, 1990.

\bibitem{c-elegans}
M.~Lechner, R.~Hasani, M.~Zimmer, T.~A. Henzinger, and R.~Grosu, ``Designing worm-inspired neural networks for interpretable robotic control,'' in {\em 2019 International Conference on Robotics and Automation (ICRA)}, (Montreal, QC, Canada), pp.~87--94, 2019.

\bibitem{ctnn}
R.~H. et~al., ``Closed-form continuous-time neural networks,'' {\em Nat. Mach. Intell.}, vol.~4, no.~11, pp.~992--1003, 2022.

\bibitem{randomforest}
T.~K. Ho, ``Random decision forests,'' in {\em Proceedings of 3rd International Conference on Document Analysis and Recognition}, pp.~278--282, 1995.

\bibitem{xgboost}
T.~Chen and C.~Guestrin, ``Xgboost: A scalable tree boosting system,'' in {\em Proceedings of the 22nd ACM SIGKDD International Conference on Knowledge Discovery and Data Mining}, (New York, NY, USA), pp.~785--794, ACM, 2016.

\bibitem{lstm}
S.~Hochreiter and J.~Schmidhuber, ``Long short-term memory,'' {\em Neural Comput.}, vol.~9, no.~8, pp.~1735--1780, 1997.

\bibitem{b1.5}
K.~Funahashi and Y.~Nakamura, ``Approximation of dynamical systems by continuous time recurrent neural networks,'' {\em Neural Netw.}, vol.~6, no.~6, pp.~801--806, 1993.

\bibitem{b2}
M.~Lechner, R.~M. Hasani, A.~Amini, T.~A. Henzinger, D.~Rus, and R.~Grosu, ``Neural circuit policies enabling auditable autonomy,'' {\em Nat. Mach. Intell.}, vol.~2, no.~10, pp.~642--652, 2020.

\bibitem{autonomousdriving}
M.~Chahine, R.~Hasani, P.~Kao, A.~Ray, R.~Shubert, and M.~L. et~al., ``Robust flight navigation out of distribution with liquid neural networks,'' {\em Sci. Robot.}, vol.~8, p.~eadc8892, Apr. 19 2023.

\bibitem{b3}
A.~Raneez and T.~Wirasingha, ``A review on breaking the limits of time series forecasting algorithms,'' in {\em 2023 IEEE 13th Annual Computing and Communication Workshop and Conference (CCWC)}, (Las Vegas, NV, USA), pp.~0482--0488, 2023.

\bibitem{sdes}
S.~Calculus, ``Course material.''
\newblock \url{https://math.nyu.edu/~goodman/teaching/StochCalc2011/SDE.pdf}.

\bibitem{b4}
M.~H. Nielsen, C.-Y. Yeh, M.~Shen, and M.~Medard, ``Blockage prediction in directional mmwave links using liquid time constant network,'' in {\em 48th International Conference on Infrared, Millimeter, and Terahertz Waves, IRMMW-THz}, 2023.

\bibitem{b5}
S.~Bothe, H.~Farooq, J.~Forgeat, and K.~Cyras, ``Time-series prediction using nature-inspired small models and curriculum learning,'' in {\em 2023 IEEE 34th Annual International Symposium on Personal, Indoor and Mobile Radio Communications (PIMRC)}, (Toronto, ON, Canada), pp.~1--6, 2023.

\bibitem{rel18}
``Technical specification 23.288. architecture enhancements for 5g system (5gs) to support network data analytics services.''
\newblock \url{https://www.3gpp.org/ftp/Specs/archive/23_series/23.288/23288-i60.zip}.

\bibitem{data}
N.~Pardhasaradhi, J.~Bose, A.~Vikram, A.~Verma, and M.~Jain, ``Identification of inefficient radios for efficient energy consumption in a mobile network,'' in {\em 2024 16th International Conference on COMmunication Systems \& NETworkS (COMSNETS)}, (Bengaluru, India), pp.~608--612, 2024.

\bibitem{KS}
J.~L.~H. Jr., ``The significance probability of the smirnov two-sample test,'' {\em Arkiv fiur Matematik}, vol.~3, no.~43, pp.~469--486, 1958.

\bibitem{ks-python}
SciPy, ``Scipy api statistical functions.'' Online.
\newblock Available: \url{https://docs.scipy.org/doc/scipy/reference/generated/scipy.stats.ks_2samp.html}.

\bibitem{addnoisetotestset}
M.~Abdelaty, R.~Doriguzzi-Corin, and D.~Siracusa, ``Aads: A noise-robust anomaly detection framework for industrial control systems,'' in {\em Information and Communications Security. ICICS 2019} (J.~Zhou, X.~Luo, Q.~Shen, and Z.~Xu, eds.), vol.~11999 of {\em Lecture Notes in Computer Science}, Cham: Springer, 2020.

\bibitem{b6}
``Kepler (kubernetes efficient power level exporter).'' Online, 2024.
\newblock Available: \url{https://github.com/sustainable-computing-io/kepler}.

\bibitem{b7}
``Exploring kepler’s potentials: Unveiling cloud application power consumption.'' Online.
\newblock Available: \url{https://www.cncf.io/blog/2023/10/11/exploring-keplers-potentials-unveiling-cloud-application-power-consu-}\newline\url{mption}.

\end{thebibliography}

%This defines the bibliographies style. Search online for a list of available styles.
\bibliographystyle{ieeetr}

\end{document}